\documentclass[11pt,a4paper]{article}
\usepackage[hyperref]{emnlp2020}
\usepackage{times}
\usepackage{latexsym}

\usepackage{microtype}

\usepackage{graphicx}
\usepackage{amsmath}
\usepackage{booktabs}

\usepackage{enumitem}
\usepackage{float} 
\usepackage{subfigure}
\usepackage{setspace}
\usepackage{helvet}  
\usepackage{courier}  
\usepackage{xcolor}
\usepackage{color}
\usepackage{multicol}
\usepackage{multirow}
\usepackage{esvect}
\usepackage{epstopdf}
\usepackage{array}
\usepackage{amsmath,amsfonts,amssymb,amsthm, bm}
\usepackage{mathrsfs}
\usepackage{verbatim}
\usepackage{enumitem}
\usepackage{latexsym}
\usepackage{setspace}
\usepackage{amsmath}
\usepackage{amssymb}
\usepackage{mathrsfs}
\aclfinalcopy 

\title{SGG: Learning to Select, Guide, and Generate \\ for Keyphrase Generation}

\author{Jing Zhao, Junwei Bao, Yifan Wang, Youzheng Wu, Xiaodong He, Bowen Zhou \\
JD AI Research \\ 
{\tt \{zhaojing857,baojunwei,wangyifan15,wuyouzheng1\}@jd.com} \\
{\tt \{xiaodong.he,bowen.zhou\}@jd.com} \\
}
\date{}

\begin{document}
\maketitle

\begin{abstract}
Keyphrases, that concisely summarize the high-level topics discussed in a document, can be categorized into \textit{present} keyphrase which explicitly appears in the source text, and \textit{absent} keyphrase which does not match any contiguous subsequence but is highly semantically related to the source.
%
Most existing keyphrase generation approaches synchronously generate present and absent keyphrases without explicitly distinguishing these two categories.
%
In this paper, a \textbf{S}elect-\textbf{G}uide-\textbf{G}enerate (SGG) approach is proposed to deal with {present} and {absent} keyphrase generation \textit{separately} with different mechanisms.
Specifically, SGG is a hierarchical neural network which consists of a pointing-based selector at low layer concentrated on present keyphrase generation, a selection-guided generator at high layer dedicated to absent keyphrase generation, and a guider in the middle to transfer information from selector to generator.
Experimental results on four keyphrase generation benchmarks demonstrate the effectiveness of our model, which significantly outperforms the strong baselines for both present and absent keyphrases generation.
%
%
Furthermore, we extend SGG to a title generation task which indicates its extensibility in natural language generation tasks.\footnote{Our code is released in \url{https://github.com/JD-AI-Research-NLP/SGG}.}
\end{abstract}

\section{Introduction\label{Sec1}}
Automatic keyphrase prediction recommends a set of representative phrases that are related to the main topics discussed in a document~\cite{liu2009}.
%
Since keyphrases can provide a high-level topic description of a document, they are beneficial for a wide range of natural language processing (NLP) tasks, such as information extraction~\cite{wan2008single}, text summarization~\cite{wang} and question generation~\cite{question}.
%
%
%


Existing methods for keyphrase prediction can be categorized into \textit{extraction} and \textit{generation} approaches. 
Specifically, keyphrase extraction methods identify important consecutive words from a given document as keyphrases,
which means that the extracted keyphrases (denoted as \textit{present keyphrases}) must exactly come from the given document. %
%
However, some keyphrases (denoted as \textit{absent keyphrases}) of a given document do not match any contiguous subsequence but are highly semantically related to the source text.
%
The extraction methods fail to predict these absent keyphrases.
Therefore, generation methods have been proposed to produce a keyphrase verbatim from a predefined vocabulary, no matter whether the generated keyphrase appears in the source text.
%
Compared with conventional extraction methods, generation methods have the ability of generating absent keyphrases as well as present keyphrases.

\begin{figure}
\centering
\setlength{\belowcaptionskip}{-7pt}
\includegraphics[width=7.6cm]{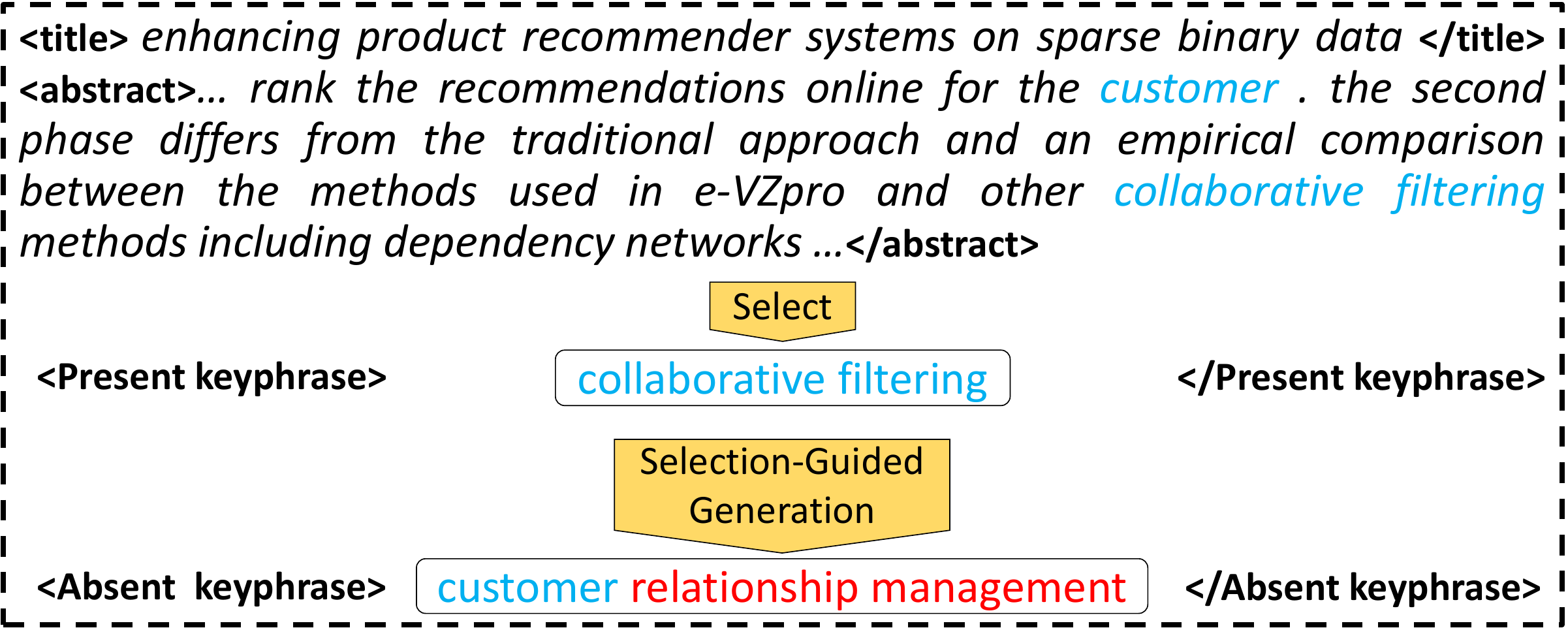}
\caption{An example of keyphrase prediction by SGG. }
\label{example}
\end{figure}

%

%
CopyRNN~\cite{meng} is the first to employ the sequence-to-sequence
(Seq2Seq) framework~\cite{sutskever2014} with the copying mechanism~\cite{gu2016} to generate keyphrases for the given documents.
%
Following the CopyRNN, several Seq2Seq-based keyphrase generation approaches have been proposed to improve the generation performance~\cite{chen2018,ye2018,chen2019guided,zhao,yue,yuan}.
All these existing methods generate present and absent keyphrases synchronously without explicitly distinguishing these two different categories of keyphrases, which leads to two problems:
(1) They complicate the identification of present keyphrases. 
Specifically, they search for words over the entire predefined vocabulary containing a vast amount of words (\textit{e.g.}, 50,000 words) to generate a present keyphrase verbatim, which is overparameterized since a present keyphrase can be simply selected from a continuous subsequence of the source text containing limited words (\textit{e.g.}, less than 400 words).
%
(2) They weaken the generation of absent keyphrases. 
%
%
%
%
Existing models for absent keyphrase generation are usually trained on datasets mixed with a large proportion of present keyphrases.
Table~\ref{absentPercentage} shows that nearly half of the training data are present keyphrases, which leads to the extremely low proportions of absent keyphrases generated by such a model, \textit{i.e.}, CopyRNN.
The above observation demonstrates that these methods are biased towards replicating words from source text for present keyphrase generation, which will inevitably affect the performance on generating absent keyphrases.
%
%

%
	
\begin{table}
\renewcommand\arraystretch{1.4}
\footnotesize
    \centering
 \begin{tabular}
 {p{1.6cm}<{\centering}|p{0.9cm}<{\centering}p{1.2cm}<{\centering}p{0.6cm}<{\centering}p{1.2cm}<{\centering}}
 
    \hline
    \hline
\multirow{2}{*}{\textbf{Training(\%)}}&
    \multicolumn{4}{c}{\textbf{Test(\%)}}\cr\cline{2-5}
    
    &\textbf{Inspec}&\textbf{Krapivin}&\textbf{NUS}&\textbf{SemEval}\\
 \hline
    49.79&13.12& 11.74&11.30&11.25\\
 \hline
 \hline
 \end{tabular}
 \caption{Proportions of absent keyphrases in training set and predictions of CopyRNN on four commonly used datasets, where top-10 predictions are considered.}
 \label{absentPercentage}
\vspace{-4mm} 
\label{bias} 
\end{table}

	
To address the aforementioned problems, we propose a \textbf{S}elect-\textbf{G}uide-\textbf{G}enerate (SGG) approach, which deals with present and absent keyphrase generation \textit{separately} with different stages based on different mechanisms. 
Figure \ref{example} illustrates an example of keyphrase prediction by SGG.
The motivation behind is to solve keyphrase generation problem from selecting to generating, and use the selected results to guide the generation.
%
Specifically, 
%
our SGG is implemented with a hierarchical neural network which performs Seq2Seq learning by applying a multi-task learning strategy.
This network consists of a selector at low layer, a generator at high layer, and a guider at middle layer for information transfer.
The selector generates present keyphrases through a pointing mechanism~\cite{vinyals2015}, which adopts attention distributions to select a sequence of words from the source text as output.
The generator further generates the absent keyphrases through a pointing-generating (PG) mechanism~\cite{see2017}.
%
%
Since present keyphrases have already been generated by the selector, they should not be generated again by the generator.
Therefore, a guider is designed to memorize the generated present keyphrases from the selector, and then fed into the attention module of the generator to constrain it to focus on generating absent keyphrases. 
%
We summarize our main contributions as follows:
\begin{itemize}[leftmargin=*]
\item We propose a SGG approach which models \textit{present} and \textit{absent} keyphrase generation separately in different stages, \textit{i.e.}, select, guide, and generate, without sacrificing the end-to-end training through back-propagation.


\item Extensive experiments are conducted to verify the effectiveness of our model, which not only improves present keyphrase generation but also dramatically boosts the performance of absent keyphrase generation. 

\item Furthermore, we adopt SGG to a title generation task, and the experiment results indicate the extensibility and effectiveness of our SGG approach on generation tasks.
\end{itemize}


\section{Related Work\label{Sec2}}
As mentioned in Section~\ref{Sec1}, the extraction and generation methods are two different research directions in the field of keyphrase prediction. 
The existing extraction methods can be broadly classified into supervised and unsupervised approaches. 
The supervised approaches treat keyphrase extraction as a binary classification task, which train the models with the features of labeled keyphrases to determine whether a candidate phrase is a keyphrase~\cite{Witten1999,medelyan2009,gollapalli2017}. In contrast, the unsupervised approaches treat keyphrase extraction as a ranking task, scoring each candidate using some different ranking metrics, such as clustering~\cite{liu2009}, or graph-based ranking~\cite{mihalcea2004,wang2014exploiting,gollapalli2014,zhang2017mike}.

This work is mainly related to keyphrase generation approaches which have demonstrated good performance on keyphrase prediction task. Following CopyRNN~\cite{meng}, several extensions have been proposed to boost the generation capability.
In CopyRNN, model training heavily relies on large amount of labeled data, which is often unavailable especially for the new domains. To address this problem, \citet{ye2018} proposed a semi-supervised keyphrase generation model that utilizes both abundant unlabeled data and limited labeled data.
CopyRNN uses the concatenation of article title and abstract as input, ignoring the leading role of the title. To address this deficiency, \citet{chen2019guided} proposed a title-guided Seq2Seq network to sufficiently utilize the already summarized information in title.
In addition, some research attempts to introduce external knowledge into keyphrase generation, such as syntactic constraints~\cite{zhao} and latent topics ~\cite{yue}.

These approaches do not consider the one-to-many relationship between the input text and target keyphrases, and thus fail to model the correlation among the multiple target keyphrases. To overcome this drawback, \citet{chen2018} incorporated the review mechanism into keyphrase generation and proposed a model CorrRNN with correlation constraints. 
Similarly, SGG separately models one-to-many relationship between the input text and present keyphrases and absent keyphrases.
%
To avoid generating duplicate
keyphrases, \citet{chen2020exclusive} proposed an exclusive hierarchical
decoding framework that includes
a hierarchical decoding process and either a
soft or a hard exclusion mechanism.
For the same purpose, our method deploys a guider to avoid the generator generating duplicate present keyphrases.
Last but most important, all these methods do not consider the difference between present and absent keyphrases.
We are the first to discriminately treat present and absent keyphrases in keyphrase generation task.

\section{Methodology\label{Sec3}}
%
%

\subsection{Problem Definition\label{definition}}
Given a dataset including $K$ data samples, where the $j$-th data item $\langle x^{(j)}, y^{(j,p)}, y^{(j,a)}\rangle$ consists of a source text $x^{(j)}$, a set of present keyphrases $y^{(j,p)}$ and a set of absent keyphrases $y^{(j,a)}$.
Different from CopyRNN~\cite{meng} splitting each data item into multiple training examples, each of which contains only one keyphrase as target, we regard each data item as one training example by concatenating its present keyphrases as one target and absent keyphrases as another one.
Specifically, assume that the $j$-th data item consists of $m$ present keyphrases $\{y^{(j,p)}_1, ..., y^{(j,p)}_m\}$ and $n$ absent keyphrases $\{y^{(j,a)}_1, ..., y^{(j,a)}_n\}$, the target present keyphrases $y^{(j,p)}$ and target absent keyphrases $y^{(j,a)}$ are represented as:
$$y^{(j,p)} = y^{(j,p)}_1 \ || \ y^{(j,p)}_2 \ || \ ...\  || \ y^{(j,p)}_m$$
$$y^{(j,a)}=y^{(j,a)}_1 \ || \ y^{(j,a)}_2 \ || \ ...\ || \ y^{(j,a)}_n$$
where $||$ is a special splitter to separate the keyphrases.
We then get the source text $x^{(j)}$, the present keyphrases $y^{(j,p)}$ and the absent keyphrases $y^{(j,a)}$ all as word sequences.
Under this setting, our model is capable of generating multiple keyphrases in one sequence as well as capturing the mutual relations between these keyphrases.
A keyphrase generation model is to learn the mapping from the source text $x^{(j)}$ to the target keyphrases $(y^{(j,p)}, y^{(j,a)})$.
For simplicity, $(x,y^p,y^a)$ is used to denote each item in the rest of this paper, where $x$ denotes a source text sequence, $y^p$ denotes its present keyphrase sequence and $y^a$ denotes its absent keyphrase sequence.

\subsection{Model Overview\label{view}}
\begin{figure}
\centering
\setlength{\abovecaptionskip}{10pt} 
\setlength{\belowcaptionskip}{-3pt}
\includegraphics[width=7.5cm]{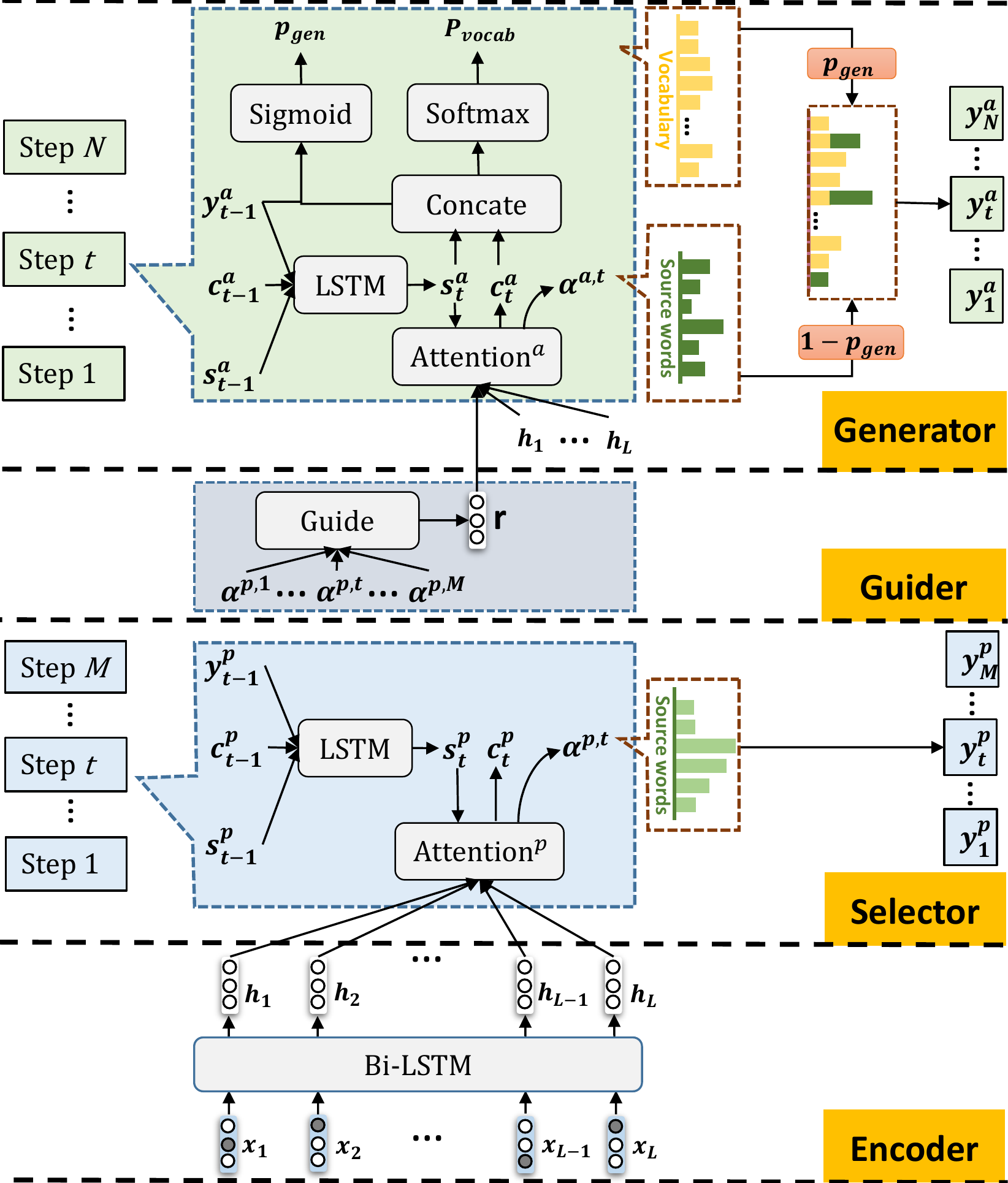}
\caption{The architecture of the proposed SGG which is implemented with a hierarchical neural network. }
\label{overview}
\end{figure}

The architecture of our proposed \textbf{S}elect-\textbf{G}uide-\textbf{G}enerate (SGG) approach is illustrated in Figure \ref{overview}. 
%
Our model is the extension of Seq2Seq framework which consists of a \textbf{text encoder}, a \textbf{selector}, a \textbf{guider}, and a \textbf{generator}.
The text encoder converts the source text $x$ into a set of hidden representation vectors $\{\mathbf{h}_i\}^L_{i=1}$ with a bi-directional Long Short-term Memory Network (bi-LSTM)~\cite{lstm}, where $L$ is the length of source text sequence.
The selector is a uni-directional LSTM, which predicts the present keyphrase sequence $y^p$ based on the attention distribution over source words.
After selecting present keyphrases, a guider is produced by a guider to memorize the prediction information of the selector, and then fed to the attention module of a generator to adjust the information it pays attention to.
The selection-guided generator is also implemented as a uni-directional LSTM, which produces the absent keyphrase sequence $y^a$ based on two distributions over predefined-vocabulary and source words, respectively.
At the same time, a soft switch gate $p_{gen}$ is employed as a trade-off between the above two distributions.

\subsection{Text Encoder\label{texEnc}}
The goal of a text encoder is to provide a series of dense representations $\{\mathbf{h}_i\}^L_{i=1}$ of the source text.
In our model, the text encoder is implemented as a
bi-LSTM~\cite{lstm} which
reads an input sequence $x$ = $\{x_i$\}$^L_{i=1}$ from two directions and outputs a sequence of forward hidden states $\{\overrightarrow{\mathbf{h}_i}$\}$^L_{i=1}$ and backward hidden states $\{\overleftarrow{\mathbf{h}_i}$\}$^L_{i=1}$ by iterating the following equations:
\begin{eqnarray}
\overrightarrow{\mathbf{h}_i} = \mathtt{LSTM}(x_i,{\mathbf{h}}_{i-1})\\
\overleftarrow{\mathbf{h}_i} = \mathtt{LSTM}(x_i,{\mathbf{h}}_{i+1})
\end{eqnarray}
The final hidden representation $\mathbf{h}_i$ of the $i$-th source word is the concatenation of forward and backward hidden states, \textit{i.e.}, $\mathbf{h}_i = [\overrightarrow{\mathbf{h}_i}; \overleftarrow{\mathbf{h}_i}]$.

\subsection{Selector\label{easyDec}}
A selector is designed to generate present keyphrase sequences through
the pointer mechanism~\cite{vinyals2015}, which adopts the attention distribution as a pointer to select words from the source text as output. 
%
%
Specifically, given source text sequence $x$ and previously generated words $\{y^p_1, ..., y^p_{t-1}\}$,
the probability distribution of predicting next word $y^p_t$ in present keyphrases is:
\begin{align}
\mathcal P(y^p_t \ | \ y^p_{<t}, x) = \mathcal \mathbf{\alpha}^{p,t}  = \mathtt{softmax}(\mathbf{u}^{p,t}) \\
u^{p,t}_i = {\mathbf{V}}^{T}_p \mathtt{tanh}({\mathbf W}_p[\mathbf{s}^p_{t};\mathbf{h}_i] + {\mathbf b}_p)
\end{align} 
where $\alpha^{p,t}$ is the attention~\cite{bahdanau2014} distribution at decoding time step $t$,
%
$i \in (1,...,L)$, and ${\mathbf{V}}_p$, ${\mathbf W}_p$ and ${\mathbf b}_p$ are trainable parameters of the model.
$\mathbf{u}^{p,t}$ can be viewed as the degree of matching between input at position $i$ and output at position $t$.
$\mathbf{s}^p_{t}$ represents the hidden state at deciding time step $t$, and is updated by equation:
\begin{equation}
 \mathbf{s}^p_{t} = \mathtt{LSTM}(y^p_{t-1},\mathbf{s}^p_{t-1},\mathbf{c}^p_{t-1})
\end{equation}
where context vector $\mathbf{c}^p_{t-1}=\sum_{i=1}^{L} \alpha^{p,t-1}_{i}\mathbf{h}_i$ is the weighted sum of source hidden states.

\subsection{Guider}
A guider is designed to fully utilize the attention information of the selector to guide the generator on absent keyphrase generation.
The idea behind is to utilize a guider $\mathbf{r}$ to softly indicate which words in source text have been generated by the selector.
This is important for helping the generator to focus on generating the absent keyphrases.
Specifically, $\mathbf{r}$ is constructed through the accumulation of the attention distributions over all decoding time steps of the selector, computed as:
\begin{equation} 
  \mathbf{r} = \sum_{t=1}^{M} \mathbf{\alpha}^{p,t}
\end{equation}
where $M$ is the length of present keyphrase sequence. $\mathbf{r}$ is an unnormalized distribution over the source words.
As the attention distribution of selector is equal to the probability distribution over the source words, $\mathbf{r}$ represents the possibility that these words have been generated by the selector.
The calculation of guider is inspired by the coverage vector~\cite{tu} that is sequentially updated during the decoding process.
In contrast to this, the guider here is a static vector which is capable of memorizing a global information.

\subsection{Selection-Guided Generator\label{hardDec}}
A generator aims to predict an absent keyphrase sequence based on the guidance of the selection information from the guider.
Unlike present keyphrases, most words in absent keyphrases do not appear in source text.
Therefore, the generator generates absent keyphrases by picking up words from both a predefined large scale vocabulary and the source text~\cite{see2017,gu2016}.
The probability distribution of predicting next word $y^a_t$ in absent keyphrases is defined as:
\begin{equation} 
\begin{aligned}
 \label{equ_8}
&\mathcal P(y^a_t\ | \ y^a_{<t}, x)\\
&  = p_{gen} \mathcal P_{vocab}(y^a_t) + (1-p_{gen}) \!\!\! \sum_{i:y^a_t=x_i}\alpha^{a,t}_{i}
\end{aligned}
\end{equation}
where $ \mathcal P_{vocab}$ is the probability distribution over the predefined vocabulary, which is zero if $y^a_t$ is an out-of-vocabulary (OOV) word. 
Similarly, if $y^a_t$ does not appear in the source text, then $\sum_{i:y^a_t=x_i}\alpha^{a,t}_{i}$ is zero.
$\mathcal P_{vocab}$ is computed as:
\begin{equation} 
 \mathcal P_{vocab}(y^a_t) = \mathtt{softmax}({\mathbf W}[\mathbf{s}^a_t;\mathbf{c}^a_t]+{\mathbf b})
\label{prob}
\end{equation}
where ${\mathbf W}$ and ${\mathbf b} $ are learnable parameters, $\mathbf{s}^a_t$ is the hidden state of generator, and $\mathbf{c}^a_t$ is the context vector for generating absent keyphrase sequence, computed by the following equations:
\begin{align} 
 \mathbf{c}^a_{t} & =  \sum_{i=1}^{L} \mathbf{\alpha}^{a,t}_{i}\mathbf{h}_i\\
 \mathcal \alpha^{a,t} & = \mathtt{ softmax}(\mathbf{u}^{a,t}) \\
 u^{a,t}_i & = {\mathbf V}^{T}_a \mathtt{tanh}({\mathbf W}_a[\mathbf{s}^a_{t};\mathbf{h}_i;\mathbf{r}] + {\mathbf b}_a)
\end{align} 
where ${\mathbf V}_a$, ${\mathbf W}_a$ and ${\mathbf b}_a$ are learnable parameters.
$\mathbf{r}$ is a vector produced by the guider.
The generation probability $p_{gen}$ at time step $t$ is computed as:
\begin{equation} 
p_{gen} = {\mathtt{\sigma}}({\mathbf W}_{gen}[\mathbf{c}^a_t;\mathbf{s}^a_t;\mathtt{emb}(y^a_{t-1})]+{\mathbf b}_{gen})
\end{equation}
where ${\mathbf W}_{gen}$ and ${\mathbf b}_{gen}$ are learnable parameters, $\mathtt{\sigma(\cdot)}$ represents a sigmoid function and $\mathtt{emb}(y^a_{t-1})$ is the embedding of $y^a_{t-1}$.
In addition, $p_{gen}$ in formula (\ref{equ_8}) is used as a soft switch to choose either generating words over vocabulary or copying words from source text based on distribution $\alpha^{a,t}$.

\subsection{Training\label{loss}}
Given the set of data pairs $\{x^{(j)}, y^{(j,p)}, y^{(j,a)}\}^K_{j=1}$, the loss function of the keyphrase generation consists of two parts of cross entropy losses: 
\begin{eqnarray}
\mathcal L_p(\theta) = -\sum_{j=1}^{K}\sum_{i=1}^{M} log(\mathcal{P}(y^{(j,p)}_i{\bf |}{ x^{(j)}};\theta)) \\
\mathcal L_a(\theta) = -\sum_{j=1}^{K}\sum_{i=1}^{N}log(\mathcal{P}(y^{(j,a)}_i{\bf |} {x^{(j)}};\theta))
\end{eqnarray}
where $\mathcal L_p$  and  $\mathcal L_a$ are the losses of generating present and absent keyphrases, respectively. $N$ is the word sequence length of absent keyphrases, and $\theta$ are the parameters in our model.
%
%
The training objective is to jointly minimize the two losses:
\begin{equation}
\mathcal L =   \mathcal L_p +  \mathcal L_a.    
\end{equation}

\section{Experiment}
\subsection{Dataset \label{dataset}}
We use the dataset collected by \citet{meng} from various online digital libraries, which contains approximately 570K samples, each of which contains a title and an abstract of a scientific publication as source text, and author-assigned keywords as target keyphrases.
We randomly select the example which contains at least one present keyphrase to construct the training set.
Then, a validation set containing 500 samples will be selected from the remaining examples.
In order to evaluate our proposed model comprehensively, we test models on four widely used public datasets from the scientific domain, namely Inspec \cite{hulth2006}, Krapivin \cite{krapivin2009}, SemEval-2010 \cite{kim} and NUS \cite{nguyen2007}, the statistic information of which are summarized in Table \ref{datatset}.

\begin{table}[H]
\setlength{\belowcaptionskip}{-10pt}
\renewcommand\arraystretch{1.4}
\footnotesize
    \centering
	\begin{tabular}{p{0.7cm}<{\centering}|p{1.2cm}<{\centering}|p{1.2cm}<{\centering}|p{1.2cm}<{\centering}|p{1.2cm}<{\centering}} 
	
	\hline
    \hline
	
	\multicolumn{2}{c|}{\textbf{Dataset}}&\textbf{\#Abs}&\textbf{\#PKPs}&\textbf{\#AKPs}\cr
	\hline
	\multirow{4}{*}{\textbf{Test}}&Inspec &500&3,654 & 1,349\cr\cline{2-5}
    &Krapivin &400& 1,299& 1,040\cr\cline{2-5}
	&NUS &211&1,333 &1,128\cr\cline{2-5}
	&SemEval & 100 & 625 & 841\cr 
    \hline
    \multicolumn{2}{c|}{\textbf{Validation}}& 500 & 1,158& 1,418\cr
    \hline
    \multicolumn{2}{c|}{\textbf{Training}}& 453,757 & 1,082,285 & 1,073,404\cr
	\hline
	\hline
	\end{tabular}
	\caption{Statistics of the dataset. \#Abs, \#PKPs, \#AKPs denote the number of abstracts, present keyphrases, and absent keyphrases, respectively.}
\label{datatset}	
\vspace{-2mm}
\end{table}

 
    

\renewcommand{\arraystretch}{1.5}
\begin{table*}[!htb]
  \centering
  \fontsize{9}{9}\selectfont
    \begin{tabular}{|c|c|c|c|c|c|c|c|c|}
    \hline
    \multirow{2}{*}{\textbf{Method}}&
    \multicolumn{2}{c|}{\textbf{Inspec}}&\multicolumn{2}{c|}{\textbf{Krapivin}}&\multicolumn{2}{c|}{\textbf{NUS}}&\multicolumn{2}{c|}{\textbf{SemEval}}\cr\cline{2-9}
    &\textbf{F1}@\textbf{5}&\textbf{F1}@\textbf{10} &\textbf{F1}@\textbf{5}&\textbf{F1}@\textbf{10} &\textbf{F1}@\textbf{5}&\textbf{F1}@\textbf{10} &\textbf{F1}@\textbf{5}&\textbf{F1}@\textbf{10}\cr
    \hline
    \hline
    \textbf{TF-IDF}&22.1&31.3&12.9&16.0&13.6&18.4&12.8&19.4\cr
    \textbf{TextRank}&22.3&28.1&18.9&16.2&19.5&19.6&17.6&18.7\cr
    \textbf{KEA}&9.8&12.6&11.0&15.2&6.9&8.4&2.5&2.6\cr
    \hline
   \textbf{CopyRNN}&27.8&\underline{34.2}&\underline{31.1}&26.6&33.4&32.6&29.3&30.4\cr
   \textbf{CopyTrans$^\dag$}&21.1&16.2&26.4&20.5&35.1&28.2&29.5&26.3\cr
    \textbf{CorrRNN}&--&--&\textbf{31.8}&\textbf{27.8}&35.8&33.0&\underline{32.0}&\underline{32.0}\cr
    \textbf{CatSeq}&\underline{29.0}&30.0&30.7&\underline{27.4}&\underline{35.9}&\underline{34.9}&30.2&30.6\cr
    \hline
    \hline
\textbf{SGG}&\textbf{30.6}&\textbf{35.9}&28.8&25.3&\textbf{36.3}&\textbf{35.8}&\textbf{33.8}&\textbf{33.6}\cr
\hline

     \hline
 \end{tabular}
  \caption{F1@5/10 results of predicting present keyphrases of different models on four datasets. The best and second best performance in each column are highlighted with bold and underline respectively. $^\dag$ indicates that the model is reimplemented.}
  \label{result_present}
\end{table*}

\renewcommand{\arraystretch}{1.5}
\begin{table}[!htb]
 
  \centering
  \fontsize{9}{9}\selectfont
    \begin{tabular}{|c|c|c|c|c|}
    \hline
    \textbf{Method}&\textbf{Inspec}&\textbf{Krapivin}&\textbf{NUS}&\textbf{SemEval}\cr
    \hline
    \hline

    \textbf{CopyRNN}&\underline{10.0}&\underline{20.2}&\underline{11.6}&\textbf{6.7}\cr
    \textbf{CopyTrans$^\dag$}&5.6&16.9&8.9&4.1\cr
    \textbf{CorrRNN$^\dag$}&8.5&15.2&8.0&3.5\cr
    \textbf{CatSeq}&2.9&7.4&3.1&2.5\cr
    \hline\hline
    \textbf{SGG}&\textbf{11.0}&\textbf{23.5}&\textbf{12.4}&\underline{4.9}\cr
\hline
 \end{tabular}
  \caption{Recall@50 results of predicting absent keyphrases of different models on four datasets.  The CorrRNN is retrained following the implementation details in~\citet{chen2018} as they did not report the Recall@50 results.}
  \label{result_absent_}
\end{table}

\subsection{Baselines and Evaluation Metrics}
For present keyphrase prediction, we compare our model with both extraction and generation approaches.
%
Extraction approaches include two unsupervised extraction methods: TF-IDF, TextRank \cite{mihalcea2004} and one classic supervised extraction method KEA \cite{Witten1999}.
%
For the generation baselines, some models, such as CopyRNN,  split each data item into multiple training examples, each of which only contains one keyphrase, while the other models concatenate all keyphrases as target. 
To simplicity, the pattern of training model only with one keyphrase is denoted as \textbf{one-to-one} and with the concatenation of all keyphrases as \textbf{one-to-many}.
The generation baselines are the following state-of-the-art encoder-decoder models:
\begin{itemize}[leftmargin=*]
\item \textbf{CopyRNN(one-to-one)}~\cite{meng} represents a 
RNN-based encoder-decoder model
incorporating the copying mechanism.
\item \textbf{CopyTrans(one-to-many)} is a transformer-based~\cite{transformer} encoder-decoder model incorporating the copying mechanism. 
\item \textbf{CorrRNN(one-to-many)}~\cite{chen2018} is an extension of CopyRNN incorporating the coverage mechanism~\cite{tu}.
\item \textbf{CatSeq(one-to-many)}~\cite{yuan} has the same model structure as CopyRNN. The difference is CatSeq is trained by one-to-many. 
\end{itemize}
The baseline CopyTrans has not been reported in existing papers and thus is retrained. The implementation of Transformer is base on open source tool OpenNMT~\footnote{
https://github.com/OpenNMT/OpenNMT-py}. 
For our experiments of absent keyphrase generation, only generation methods are chosen as baselines.
%
The copying mechanism used in all reimplemented generation models is based on the version~\cite{see2017}, which is slightly different from the implementations by version~\cite{meng,gu2016}.
\textbf{SGG} indicates the full version of our proposed model, which contains a selector, a guider, and a generator. Note that SGG is also trained under one-to-many pattern.
%

Same as CopyRNN, we adopt top-$N$ macro-averaged \textit{F-measure} (F1) and \textit{recall} as our evaluation metrics for the present and absent keyphrases respectively.
%
%
The choice of larger $N$ (\textit{i.e.}, 50 v.s. 5 and 10) for absent keyphrase is due to the fact that absent keyphrases are more difficult to be generated than present keyphrases.
%
%
For present keyphrase evaluation, exact match is used for determining whether the predictions are correct.
For absent keyphrase evaluation, Porter Stemmer is used to stem all the words in order to remove words’ suffix before comparisons.


\subsection{Implementation Details}
%
We set maximal length of source sequence as 400, 25 for target sequence of selector and generator, and 50 for the decoders of all
generation baselines.
%
%
We choose the top 50,000 frequently-occurred words as our vocabulary.
The dimension of the word embedding is 128.
The dimension of hidden state in encoder, selector and generator is 512.
The word embedding is randomly initialized and learned during training. 
We initialize the parameters of models with uniform distribution in [-0.2,0.2]. 
The model is optimized using Adagrad~\cite{adagrade} with learning rate = 0.15, initial accumulator = 0.1 and maximal gradient normalization = 2.
%
In the inference process, we use beam search to generate diverse keyphrases and the beam size is 200 same as baselines.
All the models are trained on a single Tesla P40.
\subsection{Results and Analysis \label{key}}

In this section, we present the results of present and absent keyphrase generation separately.
%
The results of predicting present keyphrases are shown in Table~\ref{result_present}, in
which the F1 at top-5 and top-10 predictions are
given.
We first compare our proposed model with the conventional keyphrase extraction methods. 
The results show that our model performs better than extraction methods with a large margin, demonstrating the potential of the Seq2Seq-based generation models in automatic keyphrase extraction task.
We then compare our model with the generation baselines, and the results indicate that our model still outperforms these baselines significantly.
The better performance of SGG illustrates the pointing based selector is sufficient and more effective to generate present keyphrase.
%

We further analyze the experimental results of absent keyphrase generation.
Table \ref{result_absent_} presents the recall results of the generation baselines and our model on four datasets.
It can be observed that our model significantly improves the performance of absent keyphrase generation, compared to the generation baselines.
%
This is because SGG is equipped with a generator that is not biased to generate present keyphrases and the designed guider in SGG further guides the generator to focus on generating absent keyphrases.
Table~\ref{SG} shows the proportion of absent keyphrases generated by SGG.
The comparison of Table~\ref{absentPercentage} and~\ref{SG} demonstrates that our model have the ability to generate large portions of absent keyphrases rather than tending to generate present keyphrases.

\begin{table}
\renewcommand\arraystretch{1.4}
\footnotesize
    \centering
	\begin{tabular}{p{1.4cm}<{\centering}|p{1.0cm}<{\centering}|p{1.2cm}<{\centering}|p{0.8cm}<{\centering}|p{1.13cm}<{\centering}} 
	\hline
	\hline
	\textbf{Method}&\textbf{Inspec}&\textbf{Krapivin}&\textbf{NUS}&\textbf{SemEval}\\
	\hline
    \textbf{CopyRNN}&13.12& 11.74&11.30&11.25\\
    \hline
	\textbf{SGG}&79.16&79.28&76.02&79.20\\
	\hline 
	\hline
	\end{tabular}
	\caption{Proportion of absent keyphrases in the predictions of CopyRNN and generator. The proportion of CopyRNN is same as Table~\ref{absentPercentage}.}
\label{SG}	
\vspace{-10mm}
\end{table}
In addition, an interesting phenomenon can be found from the results of CopyRNN and CatSeq that one-to-one pattern generally performs better than one-to-many if under the same model structure in absent keyphrase generation. 
To explore this phenomenon, we use the same code, same training set to retrain CopyRNN under one-to-one and one-to-many patterns, and the test results show that one-to-one could boost the performance in absent keyphrase generation.
However, SGG cannot be trained under one-to-one pattern as the core of guider in SGG is to memory all present keyphrases.  Even so, SGG still has better performance than CopyRNN. 
The results of SGG achieve 1.6\% average gain than CopyRNN and 31.8\% average gain than the best-performing results of one-to-many baselines over four test sets.
%

%

%

\subsection{SGG for Title Generation \label{summ}}
In this section, we explore the extensibility of SGG in other natural language generation (NLG) tasks, \textit{i.e.}, title generation.
We adopt the same dataset described in Section~\ref{dataset} for title generation, which contains abstracts, present keyphrases, absent keyphrases, and titles.
Specifically, a title generation model takes an abstract as input and generates a title as output.
To train SGG model for title generation, present keyphrases appearing in the titles are used as labels to train the selectors\footnote{The present keyphrase information used for training SGG is not used during inference. Datasets without given present keyphrases should consider to conduct labeling.}, and the titles are used to train the generators.
%
The idea behind is to utilize the present keyphrase generation as an auxiliary task to help the main title generation task.
In order to evaluate SGG on title generation, we choose models CopyTrans and pointer-generator (PG-Net)~\cite{see2017} as baselines.
%
%
We use ROUGE-1 (unigram), ROUGE-2 (bi-gram), ROUGE-L (LCS) and human evaluation as evaluation metrics.
For human evaluation, we randomly selects 100 abstracts for each test set, then distribute them to four people on average. 
The evaluation standard is the fluency of generated title and whether it correctly provides the core topics of an abstract.

\begin{table}[H]
\renewcommand\arraystretch{1.4}
\footnotesize
    \centering
	\begin{tabular}{p{1.4cm}<{\centering}|p{1.0cm}<{\centering}|p{1.0cm}<{\centering}|p{1.0cm}<{\centering}|p{1.13cm}<{\centering}} 
	\hline
	\hline
	\textbf{Inspec} &\textbf{RG-1}&\textbf{RG-2}&\textbf{RG-L}&\textbf{Human}\\
	\hline
	\textbf{CopyTrans}&83.58&43.81&45.25&74/100\\
	\textbf{PG-Net}&83.03&43.44&45.20&77/100\\
	\textbf{SGG}&\textbf{84.25}&\textbf{44.98}&\textbf{46.87}&\textbf{83/10}0\\
	\hline
	\hline
	\textbf{Krapivin} &\textbf{RG-1}&\textbf{RG-2}&\textbf{RG-L}&\textbf{Human}\\
	\hline
	\textbf{CopyTrans}&84.23&50.01&50.63&89/100\\
	\textbf{PG-Net}&84.75&50.82&51.48&87/100\\
	\textbf{SGG}&\textbf{84.96}&\textbf{51.35}&\textbf{52.34}&\textbf{90/100}\\
	\hline 
	\hline
	\textbf{NUS} &\textbf{RG-1}&\textbf{RG-2}&\textbf{RG-L}&\textbf{Human}\\
	\hline
	\textbf{CopyTrans}&86.76&\textbf{54.90}&52.49&82/100\\
	\textbf{PG-Net}&86.59&52.59&50.61&79/100\\
	\textbf{SGG}&\textbf{87.01}&\textbf{54.90}&\textbf{52.57}&\textbf{89/100}\\	
	\hline
	\hline
	\textbf{SemEval} &\textbf{RG-1}&\textbf{RG-2}&\textbf{RG-L}&\textbf{Human}\\
	\hline
	\textbf{CopyTrans}&86.92&\textbf{55.10}&53.05&82/100\\
	\textbf{PG-Net}&86.68&50.16&51.31&78/100\\
	\textbf{SGG}&\textbf{87.54}&53.38&\textbf{53.55}&\textbf{84/100}\\	
	\hline 
	\hline
	\end{tabular}
	\caption{Results of title generation of various models on four datasets.}
\label{summary}
\end{table}

\renewcommand{\arraystretch}{1.5}
\begin{table*}[!htb]
  \centering
  \fontsize{9}{9}\selectfont
    \begin{tabular}{|c|c|c|c|c|c|}
    \hline
    \multirow{2}{*}{\textbf{Dataset}}&\textbf{Absent
    keyphrase generation}&
    \multicolumn{4}{c|}{\textbf{Title generation}}\cr\cline{2-6}
    &\textbf{Recall@50}&\textbf{RG-1}&\textbf{RG-2}&\textbf{RG-L}&\textbf{BLEU-4}\cr
    \hline
    \textbf{Inspec}&8.6(-2.4)&83.51(-0.74)&44.40(-0.58)&45.80(-1.07)&11.02(+0.41)\cr\hline
    \textbf{Krapivin}&23.2(-0.3)&84.56(-0.40)&50.56(-0.79)&50.87(-0.48)&11.46(-1.38)\cr
    \hline
 \end{tabular}
  \caption{Results of SG on absent keyphrase generation and title generation tasks. ($\pm$) indicates the comparison of the results of SG and SGG. The results of SGG please refer to Table~\ref{result_absent_} and Table~\ref{summary}.}
  \label{ablation}
\end{table*}

The results of title generation are shown in Table~\ref{summary}, from which we observe that our proposed model SGG achieves better performance than the strong
baselines on all datasets, proving that SGG could be directly applied to title generation task and still keep highly effective.
%

%
%
%
%
%
\begin{figure}[H]
\centering
\setlength{\abovecaptionskip}{10pt} 
\setlength{\belowcaptionskip}{-3pt}
\includegraphics[width=7.5cm]{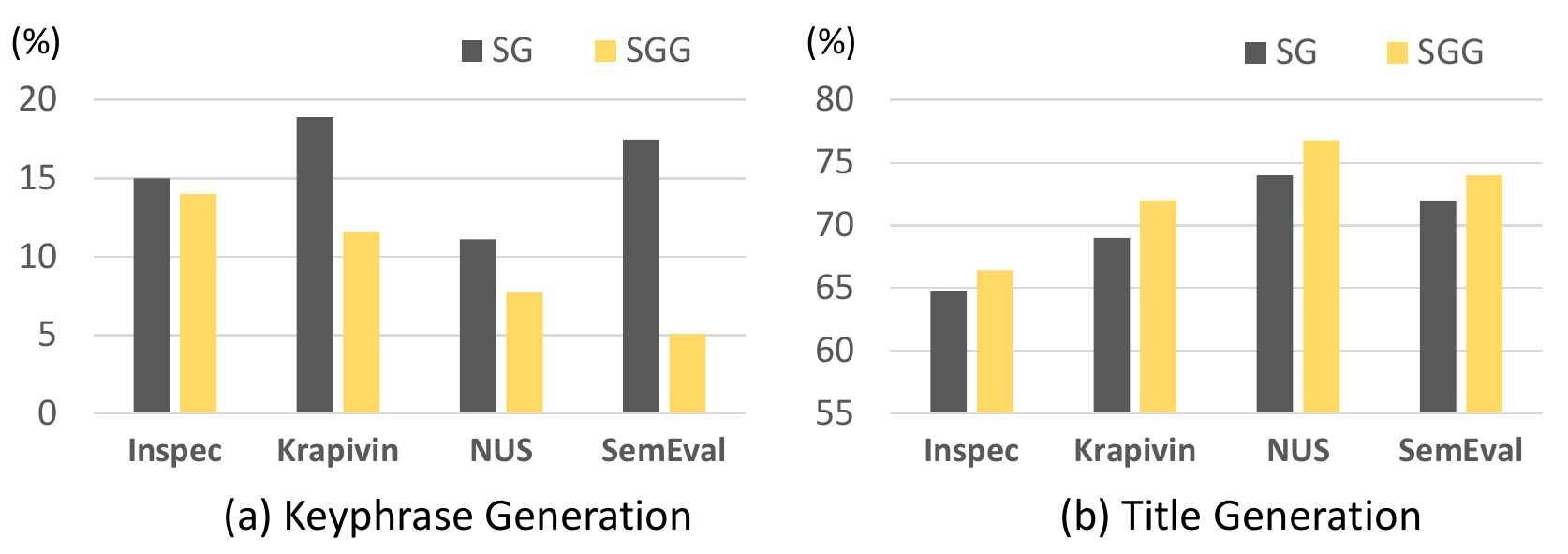}
\caption{Proportions of test examples that the predictions of generator overlap with the predictions of selector. Here only the top-1 predictions of generator and selector are used.}
\label{keyphrase}
\end{figure}
\subsection{Ablation Study on Guider \label{record}}

In this section, we further study the effectiveness of our proposed guider module. 
Table~\ref{ablation} displays the results of SG (only a \textbf{s}elector, a \textbf{g}enerator, no guider) and its comparison with SGG on the two largest test sets Inspec and Krapivin,
which illustrates that the guider has a remarkable effect on absent keyphrase and title generation tasks.
%

\begin{figure}[H]
\centering
\setlength{\abovecaptionskip}{10pt} 
\setlength{\belowcaptionskip}{-3pt}
\includegraphics[width=7.5cm]{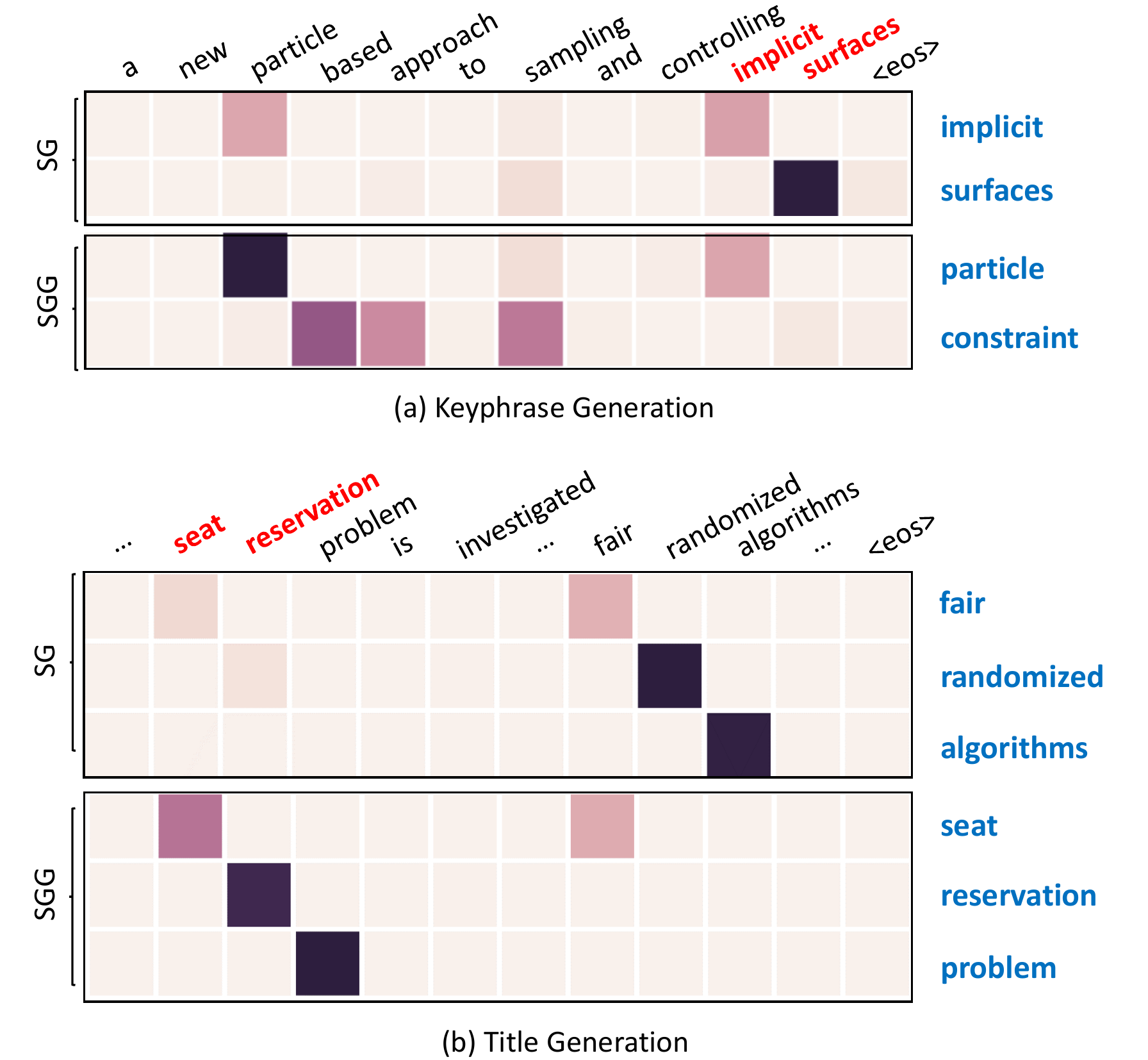}
\caption{Visualization of attention score in generator for keyphrase generation and title generation. The words marked in {\color{red} red} have already been generated by the selector. The words marked in {\color{blue} blue} are the generation of the generator. In these two examples, phrase ``particle constraint'' is the correct absent keyphrase for keyphrase generation and ``seat
 reservation problem'' is part of the correct title for title generation.}
\label{attention}
\vspace{-2mm}
\end{figure}

%
In more detail, we analyze that the function of guiders on these two tasks is different, which depends on the correlation between the targets of selector and generator.
For example, in the task of keyphrase generation, the words predicted from selector should not be repeatedly generated by generator because the present keyphrases and absent keyphrases in a given text usually do not have overlapping words.
However, in the task of title generation, the selected words by selector should be paid more attention on by generator since they are usually part of the target titles.
%
To verify the above analysis, we visualize two examples of the attention scores in generators for the two tasks in Figure \ref{attention}.
For keyphrase generation, SG repeatedly generates ``implicit surfaces'' that has already been generated by its selector.
In contrast, SGG successfully avoids this situation and it correctly generates the absent keyphrase ``particle constraint''.
For title generation, the guider helps SGG to assign higher attention scores to the words in ``seat reservation'' that has been generated by selector.

Figure~\ref{keyphrase} gives the proportion of test examples that the predictions of generator overlap with the predictions of selector.
We observe that SG is more likely to generate the words that have been generated by selector than SGG in keyphrase generation.
%
In contrast, the results on title generation indicate that SGG is more likely to generate previously selected words than SG for this task.
Through the analysis above, we conjecture that the guider is able to correctly guide the behaviour of generator in different tasks, \textit{i.e.}, learn to encourage or discourage generating previously selected words.

%
%
%

\section{Conclusion}
In this paper, a Select-Guide-Generate (SGG) approach is proposed and implemented with a hierarchical neural model for keyphrase generation, which separately deals with the generation of present and absent keyphrases.
%
%
Comprehensive empirical studies demonstrate the effectiveness of SGG.
%
Furthermore, a title generation task indicates the extensibility of SGG in other generation tasks.

\section{Acknowledgments}
This work is supported by the National Key Research and Development Program of China under Grant No. 2018YFB2100802.

\bibliographystyle{acl_natbib}
\bibliography{emnlp2020}

\end{document}